\documentclass[preprint,12pt,3p]{elsarticle} 
\usepackage[english]{babel}
\usepackage{graphicx}

\usepackage[pdftex,unicode]{hyperref}
\usepackage{url}
\usepackage{csquotes}

\usepackage{amssymb}
\usepackage{amsmath}
\usepackage[capitalise]{cleveref}

\journal{Engineering Applications of Artificial Intelligence}

\begin{document}

\begin{frontmatter}



\title{From Data to Commonsense Reasoning: The Use of Large Language Models for Explainable AI}


\author{Stefanie Krause\corref{cor}} \ead{skrause@hs-harz.de} 
\author{Frieder Stolzenburg \corref{cor}} \ead{fstolzenburg@hs-harz.de} 
\cortext[cor]{Corresponding author}

\affiliation{organization={Faculty of Automation and Computer Science, Harz University of Applied Sciences},
            addressline={Friedrichstr.~57-59},
            city={38855~Wernigerode},
            country={Germany}}


\begin{abstract} 
Commonsense reasoning is a difficult task for a computer, but a critical skill for an artificial intelligence (AI). It can enhance the explainability of AI models by enabling them to provide intuitive and human-like explanations for their decisions. This is necessary in many areas especially in question answering (QA), which is one of the most important tasks of natural language processing (NLP). Over time, a multitude of methods have emerged for solving commonsense reasoning problems such as knowledge-based approaches using formal logic or linguistic analysis. 

In this paper, we investigate the effectiveness of large language models (LLMs) on different QA tasks with a focus on their abilities in reasoning and explainability. We study three LLMs: GPT-3.5, Gemma and Llama 3. We further evaluate the LLM results by means of a questionnaire. We demonstrate the ability of LLMs to reason with commonsense as the models outperform humans on different datasets. While GPT-3.5's accuracy ranges from 56\% to 93\% on various QA benchmarks, Llama 3 achieved a mean  accuracy of 90\% on all eleven datasets. Thereby Llama 3 is outperforming humans on all datasets with an average 21\% higher accuracy over ten datasets. 

Furthermore, we can appraise that, in the sense of explainable artificial intelligence (XAI), GPT-3.5 provides good explanations for its decisions. 
Our questionnaire revealed that 66\% of participants rated GPT-3.5's explanations as either "good" or "excellent". Taken together, these findings enrich our understanding of current LLMs and pave the way for future investigations of reasoning and explainability.
\end{abstract}

\end{frontmatter}



\section{Introduction}\label{intro} 
Commonsense knowledge encompasses everyday assumptions about the world \cite{zang2013survey}. It is typically acquired through personal experience but can also be inferred by generalizing from common knowledge \cite{speer2008analogyspace}. 
Humans employ a diverse range of knowledge and reasoning to comprehend the meanings of language. \cite{storks2019recent}. 
Commonsense reasoning enables us to link various pieces of knowledge to draw new conclusions \cite{storks2019recent}. 
For instance, consider the sentence from the commonsense reasoning dataset CODAH \cite{CD+19}, "Sally thought the test was a piece of cake." It is easy for us to infer that Sally found the test easy. While this kind of knowledge and reasoning is instinctive for human readers, it remains challenging for machines \cite{storks2019recent}. 

LLMs play a crucial role in developing adaptable, general language systems \cite{BM+20}. 
Lately, a media hype was triggered by the LLM ChatGPT.\footnote{\url{https://chat.openai.com/}} This model from OpenAI provides a user-friendly interface and excels across various tasks, e.g., businesses can rapidly and accurately address customer inquiries, thereby freeing up resources and delivering a more personalized customer experience \cite{DL23}. 
In the education sector, students use this new technology for assignment writing and exam preparation among other tasks~\cite{krause2024evolution}. In healthcare, LLMs can be used, e.g., for effective medical documentation \cite{goyal2024healai}, in legal departments as tax attorney \cite{nay2024large} or in combination with robotics \cite{wang2024large, zeng2023large}. 
To facilitate the application of AI across diverse domains, the need for explainability in these systems is becoming more crucial. 
According to \cite{BarredoArrieta.2020}, it is essential to consider the target audience when seeking explainability. They define XAI as follows:
\enquote{Given an audience, an XAI is one that produces details or reasons to
	make its functioning clear or easy to understand.} 
Explainability refers to presenting a model's outcome in a human-readable format, such as through explanatory text. 
In our work, we focus on explainability of AI models in the above sense with the goal of XAI to provide human-readable explanations to make users understand the automated decision-making of large language models a posteriori.

There is a strong link between XAI and commonsense reasoning, as both focus on enhancing the explainability of AI models. Commonsense reasoning can enhance the explainability of AI models by enabling them to provide intuitive and human-like explanations for their decisions. 
According to \cite{Davis.2015}, starting with a better understanding of human cognition is a solid foundation. Humans use cognitive reasoning to draw meaningful conclusions despite incomplete and inconsistent knowledge \cite{FH+19a}. For us, cognitive reasoning is particularly useful when we encounter new situations that are not covered by formal rules or guidelines. 
In these situations, we rely on our commonsense to make judgments and decisions that are appropriate and effective. Furthermore, commonsense reasoning is essential in interpersonal interactions and communication, as it allows us to understand the perspectives of others and to navigate social situations effectively.
Commonsense reasoning can help AI models to be more robust in the context of novel situations. A model that can reason based on commonsense principles is better equipped to handle situations that it has not explicitly encountered before, as it can draw on its general understanding of the world to make informed decisions. So far, commonsense reasoning is intuitive for humans but has been a long-term challenge for AI models.

We assume that an LLM can reason similar to humans without the need of logical formulas or explicit ontology knowledge. 
Recent advances in LLMs (e.g. \cite{Liu.26.07.2019}) have pushed machines closer to human-like understanding capabilities. 
We believe that language comprehension and commonsense reasoning do not require formal structures, although they eventually may provide a better understanding afterwards for humans. Instead, we assume that LLMs are the appropriate way towards human-like ability to reason as well as explain AI decisions.
To tackle the growing demand for explainable AI systems we aim to show that LLM generated explanations are helpful for users to understand AI decisions. There is no specific structure of learning necessary: LLMs can generate human-like explanations a posteriori. For this reason, we formulate the following \textbf{hypotheses}:

\begin{enumerate}
	\item LLMs can handle commonsense reasoning in question answering tasks with near-human-level performance.
	\item LLMs are able to generate good, human-understandable explanations for their decisions.
\end{enumerate}

We continue our paper by giving an overview of important research directions in \cref{found}. Then, we evaluate the performance of the recent LLMs GPT-3.5, Gemma and Llama 3 on commonsense reasoning tasks in \cref{eval1}. Since measuring explainability is still a problem we address this by first testing the LLMs on eleven QA datasets where commonsense capabilities are required. Using a random sample of each benchmark dataset, we subsequently evaluate the quality of GPT-3.5's responses with a questionnaire (\cref{eval}).

The main \textbf{contributions} of this paper are described in \cref{futur} and can be summarized as following:
\begin{itemize}
	\item evaluation of GPT-3.5's, Gemmas's and Llama 3's ability to perform commonsense reasoning
	\item quality measurement of GPT-3.5's explanations by a questionnaire
\end{itemize}

\section{Related Work}\label{found}

\subsection{Commonsense Reasoning Approaches}
Commonsense reasoning is a difficult task for a computer to handle \cite{SSS19b}. To address this problem, various approaches have been followed in the past. 
McCarthy~\cite{mccarthy1959programs} was the first who outlined the basic approach of representing commonsense knowledge with predicate logic. 
Symbolic logic approaches were the main representation type, see e.g. \cite{Forbus.1984,Lenat.1995}. While still in use today \cite{Davis.2017} for this extremely complex task to work well it requires a large amount of additional logical scaffolding to precisely define the terms used in the statement and their  interrelationships \cite{Liu.2004}.

There is a big gap between the logical approach with deductive reasoning and human reasoning, which is largely inductive, associative, and empirical, i.e., based on former experience. Human reasoning, in contrast to formal logical reasoning, does not strictly follow the rules of classical logic. There have been efforts to utilize an approach which uses an automatic theorem prover (that allows to derive new knowledge in an explainable way), large existing ontologies with background knowledge, and recurrent networks with long short-term memory (LSTM) \cite{Hochreiter.1997} but still did not stand out much from the baseline \cite{SSS19b}.

Recent efforts to acquire and represent commonsense knowledge resulted in large knowledge graphs, acquired through extractive methods \cite{Speer.2017} or crowdsourcing \cite{Sap.2019b}. Some approaches use supervised training on a particular knowledge base, e.g., ConceptNet for commonsense knowledge. ConceptNet is a crowd-sourced database that represents commonsense knowledge as a graph of concepts connected by relations \cite{Speer.2017}. 

Interestingly, LLMs (cf. \cref{llms}) do not contain any explicit semantic knowledge or grammatical let alone logical rules that would allow an explicit reasoning process, not even the large ontologies from the logical knowledge representation like Cyc \cite{lenat1995cyc} or Adimen-SUMO \cite{alvez2012adimen}. A way out might be to have neural networks learn reasoning explicitly, possibly by focusing on certain sentence forms as in syllogistic reasoning maybe implemented with neural-symbolic cognitive reasoning by specifically structured neural networks \cite{ABG01,ALG09,ZF+21}. In contrast to simple deep learning, information from different places and/or documents must be merged here in any case. It does not suffice to investigate any local text properties, e.g., determining the text form.

There are different types of commonsense reasoning, e.g., causal, temporal, physical,  social etc. (see \cref{datasets}), each with its unique characteristics and applications. Understanding how these different reasoning tasks can be solved with LLMs is essential. 

\subsection{LLMs}\label{llms} 
In the past, most deep learning methods used supervised learning and therefore required substantial amounts of manually labelled data. Recent research has shown that learning good representations in an unsupervised fashion can provide a significant performance boost. 
The capacity of LLMs is essential to the success of zero-shot task transfer \cite{RW+19}. 
An example of a premier LLM that can handle a wide range of natural language processing tasks is Open\-AI’s GPT-3 \cite{BM+20}. GPT-3 (Generative Pre-trained Transformer) is a third-generation, autoregressive language model that uses unsupervised learning to produce human-like text.  In our further investigation, we will focus on version GPT-3.5 with 175B parameters, as this chatbot based model is regarded as one of the most groundbreaking LLMs.

We further utilise the open-source model Meta-Llama-3-70B-Instruct by Meta. This is a instruction-tuned version with 70 billion parameters \cite{touvron2023llama, meta2023llama}. Nevertheless, we use the very lightweight model with only 7 billion parameters "Gemma-1.1-7b-it" by Google. This model is open-source and built for responsible AI development based on the same research and technology used to create Gemini models \cite{team2024gemma}. 
We carefully chose these 3 LLMs, as they are state-of-the-art, differ in size (see \cref{tab:LLMsize}) and are mostly open-source.
\begin{table} [h!]
	\begin{center}
		\caption{LLM size comparison with respect to number of parameters/neurons in unit B = billion.}\label{tab:LLMsize}
		\begin{tabular}{c c}
			\hline
			\textbf{Model} & \textbf{billion parameters} \\
			GPT-3.5 & 175B \\
			Llama 3 & 70B \\
			Gemma & 7B \\
			\hline
		\end{tabular} 
	\end{center}
\vspace{-0.5cm}
\end{table}

\subsection{Combining LLMs and Reasoning}
Reasoning is one of the most actively discussed and debated capabilities of LLMs. 
However, it is important to note that proficiency in language does not necessarily equate to strong reasoning abilities in models \cite{mahowald2024dissociating}. 
Experts continue to debate the ability of LLMs to reason effectively in zero-shot scenarios  \cite{espejel2023gpt}. 
While some researchers argue that LLMs demonstrate satisfactory zero-shot reasoning capabilities \cite{kojima2022large}, others contend that the models struggle with planning and reasoning tasks \cite{valmeekam2022large}. 

ChatGPT, for example, is often seen as an unreliable reasoner due to issues such as hallucinations, a problem common to many LLMs \cite{bang2023multitask}. 
Specifically, ChatGPT-3.5 faces significant challenges in certain areas, such as mathematics \cite{espejel2023gpt}. The mathematical performance of both GPT-3.5 and GPT-4 falls well below the level of a graduate student \cite{frieder2024mathematical}. Attempts to combine elementary mathematics with commonsense reasoning have shown that no existing AI systems can reliably solve these problems \cite{davis2024mathematics}.
Thus, we investigate the reasoning ability of different current LLMs in a fine-grained manner, which includes causal, temporal, comparative, physical, social, numerical reasoning, etc., via question-answering tasks. 

\section{Evaluating LLMs on QA Tasks} \label{eval1}
We assess the LLMs twofold: First, we evaluate the accuracy of GPT-3.5, Gemma and Llama 3 on QA benchmarks with multiple-choice questions. 
Please note that the analysis of GPT-3.5 has been taken from our previous work in 2023 \cite{krause2023commonsense}. The evaluation of Gemma and Llama 3 was conducted in May 2024. 
In the commonsense reasoning benchmarks we considered, the correct answer is indicated, although it is not always clear whether this answer really is objectively the best one. 
Second, we use part of the questions from the QA benchmarks for a questionnaire to evaluate the quality of all three LLM responses compared to human accuracy. Furthermore, we evaluate the explanation quality of GPT-3.5 with the help of our questionnaire. 

\subsection{Benchmark Datasets} \label{datasets}
We utilize 11 publicly available benchmark datasets carefully designed to be difficult to solve without commonsense knowledge. From each dataset, we select 30 random examples, covering different QA tasks like text completion, understanding of cause and effect or temporal relationships. In addition, different fields like medicine, physics, and everyday life situations are covered. We evaluate the performance of the LLMs with the QA benchmarks presented in \cref{tab:datasets}.
\begin{table*}[h!]
	\begin{center}
		\caption{Overview of 11 datasets for commonsense reasoning. For each dataset we report the year the dataset was published, the number of choices each QA task in the respective benchmark has, and the knowledge domain or reasoning scenario.}\label{tab:datasets}
		
		\begin{tabular}{llcll}
			\hline
			\textbf{dataset} & \textbf{year} & \textbf{choices} & \textbf{knowledge domain or reasoning scenario} \\
			Story Cloze Test \cite{storyclozetest}& 2017 & 2 & causal, temporal \\
			CREAK \cite{Onoe.03.09.2021} & 2021 & 2 & physical, social\\ 
			CODAH \cite{CD+19} & 2019 & 4 & situations observed in videos \\
			COM2SENSE \cite{Singh.2021} & 2021 & 2 & causal, comparative, temporal, social, numerical \\ 
			CosmosQA \cite{Huang.2019} & 2019 & 4 & causal \\ 
			e-CARE \cite{DuLi.2022} & 2022 & 2 & causal \\
			ARC \cite{Clark.2018}& 2018 & 4 & natural, grade-school science\\
			Social~IQa \cite{Sap.2019} & 2019 & 3 & social, comparative\\
			COPA \cite{MR11} & 2011 & 2 & causal\\
			MedMCQA \cite{Pal22} & 2022 & 4 & medical subjects \\
			CommonsenseQA \cite{Talmor.02.11.2018} & 2018 & 5 & semantic relations\\
			\hline
		\end{tabular}
	\end{center}
\end{table*}

\subsection{LLM Results on Different Datasets} \label{method}
\begin{figure*}[h!]
	\centering
	\includegraphics[scale=0.95]{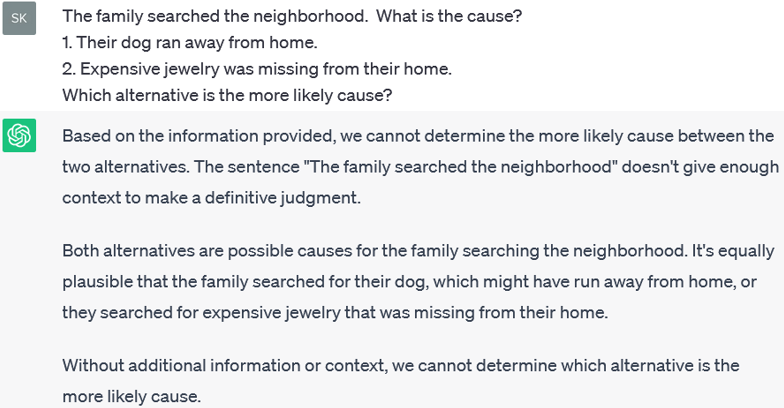}
	\caption{Example for a invalid response from GPT-3.5 due to insufficient context information (COPA example 612).} 
	\label{fig:notvalid}
\end{figure*}

Using 30 randomly selected examples from each dataset we tested the respective QA tasks with GPT-3.5, Gemma and Llama~3. We analysed the accuracy and further differentiate between incorrect and invalid answers. Invalid means that the LLM does not respond which answer option is correct and instead asks for further context information, see \cref{fig:notvalid} for an example. 
However, invalid answers have only been an issue with the COPA dataset and the models GPT-3.5  and Llama 3. GPT-3.5 (33.33\% invalid responses) has a lot more troubles when short context is provided than Llama 3 (3.33\% invalid responses).

\cref{tab:accQAtaks} shows the accuracy of the three LLMs over all datasets. GPT-3.5 and Gemma performs overall on a similar level with around 72\% overall accuracy, outperformed by Llama 3 with nearly 90\% accuracy. 
\begin{table*} 
	\begin{center}
		\caption{Results for 11 datasets of commonsense reasoning. For each dataset we report the accuracy of GPT-3.5, Gemma and Llama 3 on 30 randomly selected examples per dataset.}\label{tab:accQAtaks}
		
		\begin{tabular}{lcrr}
			\hline
			\rule{0pt}{12pt}%
			\textbf{dataset} & \textbf{GPT-3.5} & \textbf{Gemma} & \textbf{Llama 3}\\
			Story Cloze Test \cite{storyclozetest} & 93.33\% & 83.33\% & 100.00\%\\
			CREAK \cite{Onoe.03.09.2021} & 86.67\% &86.67\% &93.33\%\\
			CODAH \cite{CD+19} & 80.00\% &66.67\% &97.67\%\\
			COM2SENSE \cite{Singh.2021} & 76.67\% &76.67\% &93.33\%\\
			CosmosQA \cite{Huang.2019} & 76.67\% &66.67\% &100.00\%\\
			e-CARE \cite{DuLi.2022} & 76.67\% &66.67\% &80.00\%\\
			ARC \cite{Clark.2018} & 70.00\% &70.00\% &93.33\%\\
			Social~IQa \cite{Sap.2019}  & 66.67\% &73.33\% &83.33\%\\
			COPA \cite{MR11} & 63.33\% &83.33\% &93.33\%\\
			MedMCQA \cite{Pal22} & 60.00\% &60.00\% &93.33\%\\
			CommonsenseQA \cite{Talmor.02.11.2018} & 56.67\% &46.67\% &60.00\%\\
			\hline
			\hline
			overall & 73.33\% & 70.91\% & 89.70\%\\
			\hline
		\end{tabular}
	\end{center}
\end{table*}
We found that all models have the lowest performance on CommonsenseQA dataset with less then 60\% accuracy. The CommonsenseQA dataset has with 5 the most answer options and uses semantic relations.  
The medical dataset MedMCQA was very difficult for GPT-3.5 and Gemma (both 60\% accuracy), however, Llama 3 did perform quite impressive on this dataset with 93\% accuracy. Similar results for the ARC dataset, where GPT-3.5 and Gemma both achieved 70\% accuracy, while Llama 3 outperforms the models with 93\% accurate answers. 
The best performance was achieved by Llama 3 with 100\% accuracy on the CosmosQA and the Story Cloze Test. For the Story Cloze Test dataset GPT-3.5 achieved its best results with 93\% accuracy and Gemma reaching 83\% accuracy.
Before we start with the analysis of the results, we should acknowledge that we did not take into account the stochastic elements of LLMs, wherein repeated queries might result in varying answers.

\subsection{Analysis of LLM Results}\label{anal}
In our error analysis, we identified seven types of challenges where some LLMs still face problems:
\begin{enumerate}
	\item \textbf{missing context}: When GPT-3.5 lacks sufficient context, it may occasionally be unable to provide an answer to the QA task. This has happened 10 times in total and solely with examples of the COPA dataset. This could be due to the very short premise texts in the COPA dataset, see \cref{fig:notvalid}. In this dataset, the premise texts consist of only five to nine words (on average six words) in the cases where GPT-3.5 complained about not having enough information to answer the question. In some of those cases, GPT-3.5 reveals which context information is missing:
	\enquote{The actual outcome would depend on a variety of factors, such as the political climate, the credibility of the politician, and the specific details of the argument in question. Without this information it is impossible to determine which alternative is more likely.} (COPA example 619).
	
	\item \textbf{comparative reasoning}: Both GPT-3.5 and Gemma have problems when more than one answer option is plausible. This is the case in comparative scenarios in the COM2SENSE and Social~IQa dataset. In such cases, the commonsense reasoner must explicitly investigate the likelihood of different answer candidates. For the Social~IQa example 26823 \enquote{Sasha was throwing a party in her new condo which they bought a month ago. What does Sasha need to do before this?} GPT-3.5 answers \enquote{Turn music on} which is likely but the correct and even more likely answer is \enquote{needed to buy food for the party}.
	
	\item \textbf{subjective reasoning}: Some answers depend on the personality of the reasoner, e.g. Social~IQa example 18571: \enquote{Alex's powers were not as strong since he was just starting out. Alex used Bailey's powers since hers were stronger. How would Bailey feel as a result?} the correct answer according to the benchmark is \enquote{good} but instead GPT-3.5 answers \enquote{upset} and explains \enquote{Bailey may feel that her powers are being taken advantage of \dots} which we think is more a personalized subjective inference instead of a commonsense answer.
	
	\item \textbf{slang, unofficial abbreviations, and youth language}: GPT-3.5 and Gemma have difficulties to understand slang, unofficial abbreviations and youth language like \enquote{subs} for \enquote{subscribers} or \enquote{yrs} for \enquote{years}. This could be observed, e.g., in Cosmos~QA examples 6599 and 5748.
	
	\item \textbf{social situations}: 
	We identified a particular difficulty in correctly understanding social situations in the Social~IQa dataset. For instance, when asked the question \enquote{Kai was visiting from out of state and brought gifts for Quinn's family. What will Kai want to do next?} GPT-3.5 choose the answer \enquote{needed to leave his hometown} instead of the correct answer option \enquote{watch the opening of gifts} (Social~IQa example 6863).
	
	\item \textbf{medical and science domain}: The analysis of MedMCQA showed that GPT-3.5 and Gemma is lacking a deep domain knowledge in the medical field. The answers of GPT-3.5 were always plausible and explained with a lot of details (on average 43 words per explanation) but 40\% were incorrect. This is assumed to be because of many medical technical terms that are not common knowledge, e.g., \enquote{Styloglossus muscle} or \enquote{Genioglossus muscle} that are different muscles in the tongue (MedMCQA example 23b363d6-8210-4657-b293-54c9e28bdf31). For non-medical professionals, these questions are difficult to answer, too. Furthermore, the natural, grade-school science questions in the ARC dataset achieved a rather low accuracy of 70\% with both GPT-3.5 and Gemma.
	
	\item \textbf{semantic relations}: All LLMs had their worst performance on the CommonsenseQA dataset with five answer options and usage of multiple target concepts that have the same semantic relation to a single source concept. This indicates that semantic relations are difficult to understand for the LLMs.
\end{enumerate}

Please be aware that for certain questions to be answered correctly, one must possess in-depth knowledge rather than commonsense reasoning ability, e.g., you have to know that \enquote{Prison Break} is a television show, not a movie in a theatre to tell that \enquote{The couple went to the movie theatre to watch Prison Break} is a wrong statement (CREAK example~98). 

\section{Questionnaire}\label{eval}
\subsection{Design of the Questionnaire}
To further evaluate the quality of LLM responses on different benchmark datasets and to make a comparison to human performance, we created a questionnaire in our previous work \cite{krause2023commonsense}. Since the number of participants was quite low, we conducted the questionnaire again intending to reach more participants. For a detailed description of the questionnaire we refer to our previous work \cite{krause2023commonsense}. We used two randomly selected examples for each of the above mentioned datasets -- except for MedMCQA, as we believe these questions are too challenging for individuals without a medical background to answer.

\begin{figure}[h!]
	\centering
	\includegraphics[scale=0.8]{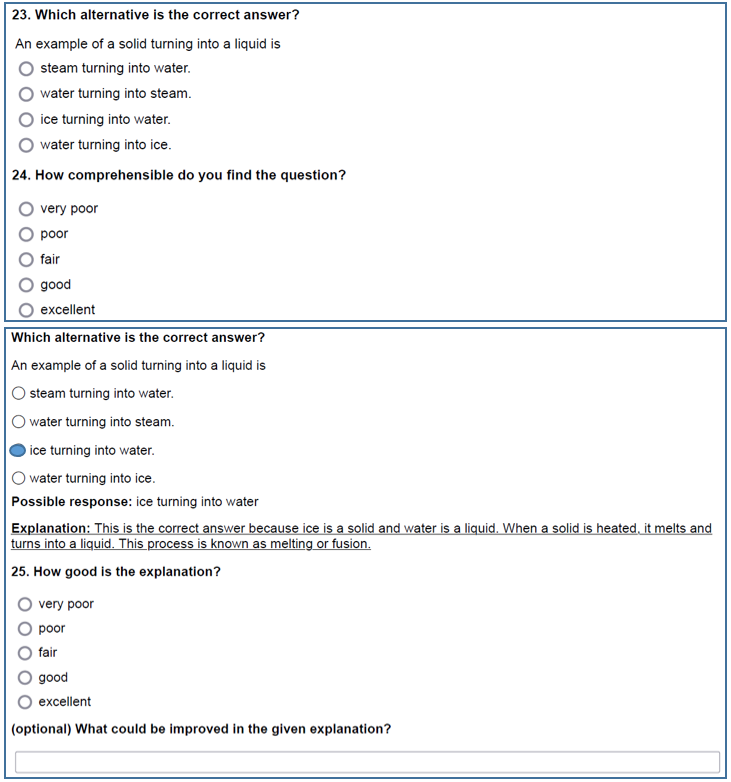}
	\caption{Example of one QA task with the common structure (ARC example). Only after participants answer the first two questions the next two questions with the possible response and explanation are shown.}
	\label{fig:q1}
\end{figure}

We developed a publicly open online questionnaire. Participation was voluntary; participants could not be identified from the material presented and no plausible harm to participating individuals could arise from the study. Survey content validity was reviewed in a pretest by one professor, one academic staff and one non-academic volunteer (business consultant) who did not participate in developing the survey. The questionnaire was structured in three parts, the first containing demographic and personal information (gender, age, nationality, English level). The main part then consists of the QA tasks of the different datasets as well an evaluation of GPT-3.5 explanations. Note that the survey participants did not know that \emph{all} explanations have been generated by GPT-3.5. 
An example of the main questionnaire section is provided in \cref{fig:q1}.  
We used this structure for 10 datasets and randomly selected two examples from every dataset. Therefore we considered $2 \cdot 10 = 20$ QA tasks in our questionnaire. 
	
\subsection{Questionnaire Participants}
	In total, 152 people participated in the questionnaire, but because of missing data, we only used the responses of 81 participants. The time to fully answer the whole questionnaire was about 25 minutes, which is probably why several participants did not complete the questionnaire until the last question.
	The participant's English level was mainly advanced or excellent so there was no language barrier in understanding the QA tasks. Among the completed questionnaires, 
	the average age was rather young with 26 years, with a minimum of 19 years and maximum of 51 years. Most of the participants were from India or Germany and the rest was from Bangladesh, Pakistan, Finland, Russia, Turkey, Canada, Iran, Lebanon and Switzerland.
	
\subsection{Questionnaire Responses}
We found that the participants answered with a mean accuracy of 69.00\% on a subset of QA tasks compared to GPT-3.5 74.33\%, Gemma 72.00\% and Llama 3 89.00\% (mean accuracy over all datasets, except MedMCQA). 
Note that 20 in detail analyzed QA tasks of the questionnaire are not as representative as the over 300 QA tasks from \cref{method}, even though we selected the 20 QA tasks randomly. 

\begin{figure*}[h!]
	\centering
	\includegraphics[scale=0.9]{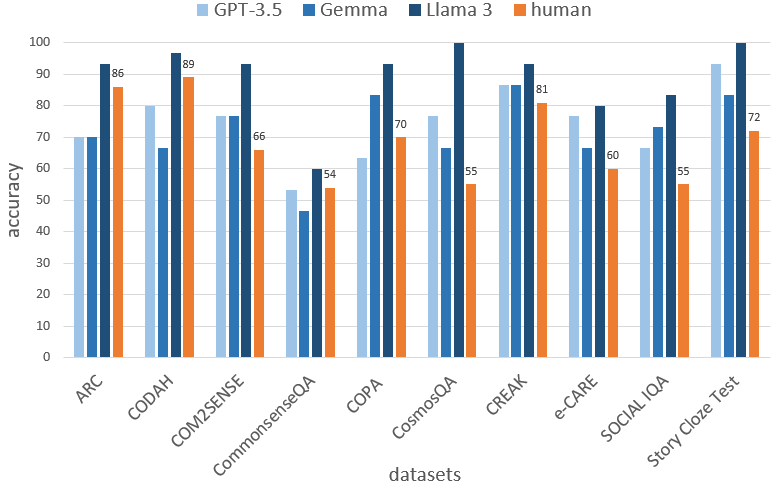}
	\caption{Comparison of accuracy of human questionnaire participants (organge) and three different LLMs: GPT-3.5, Gemma and Llama 3 (blue) on ten different datasets.}
	\label{fig:humanvsChatGPT}
\end{figure*}
A comparison of the performance of three different LLMs to the questionnaire participants on the different datasets is presented in \cref{fig:humanvsChatGPT}. One can recognize very quickly that Llama 3 outperforms human accuracy on every dataset. We found that Llama 3 is outperforming humans impressively by 21\% more accuracy on average over ten datasets. GPT-3.5 and Gemma outperformed humans on at least six datasets. On the CommonsenseQA dataset the results have been very similar. 
The dataset the LLMs perform worst is also the same dataset humans have the most problems (CommonsenseQA). The greatest difference between human and LLM performance is on the CosmosQA dataset, where Llama 3 outperforms the survey participants by 45\% more accuracy. 
When comparing human and GPT-3.5 performance the greatest difference is on the Story Cloze Test and Cosmos~QA datasets. In these two datasets, GPT-3.5 outperforms humans with over 20\% difference in accuracy. 
It is quite interesting that all LLMs perform better on Cosmos~QA than the survey participants as contextual commonsense reasoning is needed for this dataset. It focuses on reading between the lines over a diverse collection of people's everyday narratives. 
In contrast, humans perform a lot better than GPT-3.5 and Gemma on ARC where an understanding of science questions is needed. 

We want to make sure that we recognize any correlations regarding the tasks comprehensibility influencing the accuracy of answers or the explanation quality. 
We found that 82\% of the questions used in our questionnaire were rated "excellent", "good" or "fair"  comprehensible. The SocialIAQ examples have been rated worst by our human participants. The human accuracy on this QA task is only 55\%, however GPT-3.5 reached 67\% accuracy. 
We aimed to explore the relationship between the comprehensibility of tasks and GPT-3.5's explanations. It is worth noting that most questions of the different QA tasks are comprehensible according to the participants. We observed that there is a mean linear positive correlation of 0.64 between the comprehensibility of the QA tasks and that of GPT-3.5's explanations. This means that the way the users comprehend the QA tasks has an impact on the estimated quality of the explanation from GPT-3.5. 

\begin{figure*}[h!]
	\centering
	\includegraphics[scale=0.6]{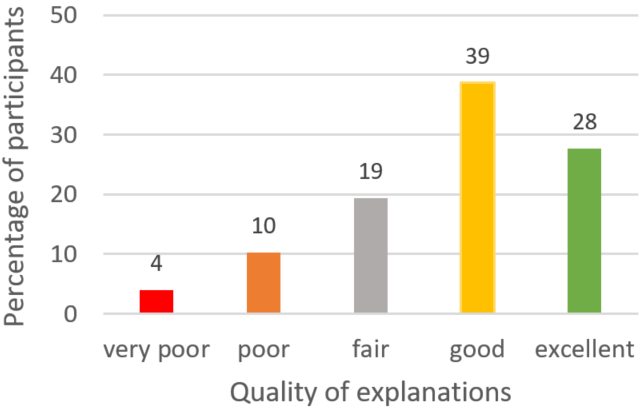}
	\caption{Participants’ rating of GPT-3.5's explanation quality on a 5 Likert scale from \enquote{very poor} to \enquote{excellent}.}
	\label{fig:explanation}
	\vspace{-0.5cm}
\end{figure*}
The quality of GPT-3,5's explanations is visualised in \cref{fig:explanation}. In general, explanations were mostly rated \enquote{good} or \enquote{excellent} with 66\% and only 42 times very poor. Explanations were rated \enquote{fair} or better with 86\%. 
Furthermore, we found that the explanations for COPA example 610 were often rated \enquote{poor} or \enquote{very poor}, and for this example GPT-3.5's answer was invalid as it could not decide for one option saying: \enquote{It is not specified in the given information which alternative is the more likely cause.} 
The average length of GPT-3.5's explanations is 38 words for both correct and incorrect responses. From the optional free text field, we received mostly the same possible improvement for GPT-3.5's explanation: GPT-3.5 should explain why the other answering options are false or less likely and not only focus on explaining why one option is correct. This is in particular important in comparative reasoning tasks. Further, the explanation should provide more information than stated in the question, however, the sentences should not be too long.

\begin{figure*}[h]
	\centering
	\includegraphics[scale=0.4]{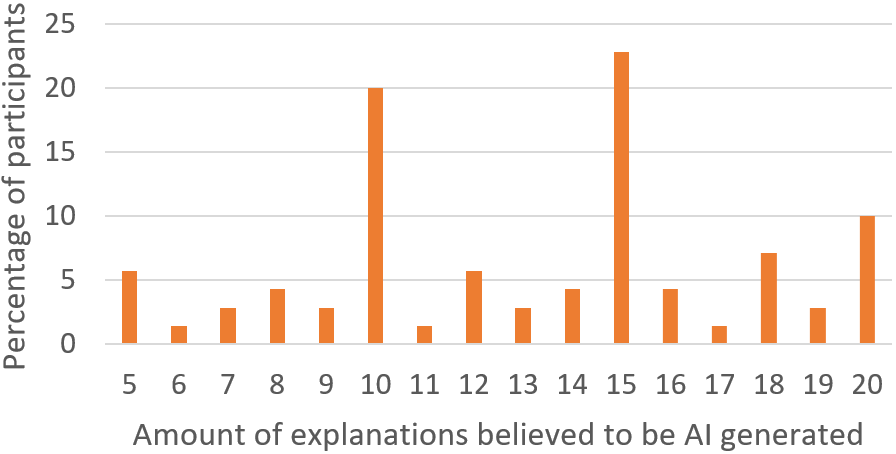}
	\caption{Participants believe how many explanations from zero to twenty are generated by an AI. Actually all 20 explanations were generated by GPT-3.5. 
	}
	\label{fig:numberAIexplanations}
\end{figure*}

\cref{fig:numberAIexplanations} shows the survey participants' guess of how many responses are created by an AI. In the chart, one can see that the mode is 15 explanations. While all respondents thinks that at least five explanations are generated by an AI, the mean amount of AI explanations is 13. Thus all participants believe that at least 25\% of the explanations were AI generated and most believe that more than half of the explanations are AI generated. In fact, all explanations are generated with GPT-3.5.

\begin{figure*}
	\centering
	\includegraphics[scale=0.8]{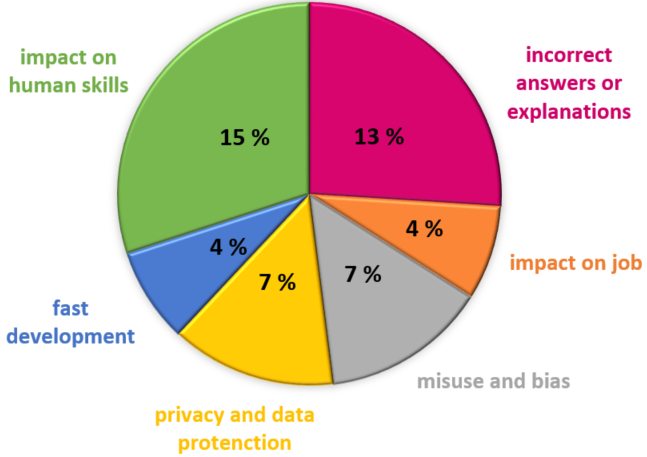}
	\caption{Concerns regarding the use of AI tools mentioned by questionnaire participants have been categorized into six distinct groups.}
	\label{fig:concerns}
\end{figure*}
We found that most study participants are not very concerned about using AI (chatbot) tools, the largest number of participants are "somewhat concerned" with 41.98\% and even 17.28\% are "not concerned at all", while only 5 participants are "very concerned". However, 37 participants expressed multiple concerns in a free-text field, which we subsequently organized into six clusters. The results are visualized in \cref{fig:concerns}. 

Most often the negative impact on human skills like critical thinking, creativity, cheating in education, loss of confidence in own abilities or dependence on AI tools is mentioned.  
The second important category of concern is incorrect answers or explanations. Many participants stated that LLMs can generate false or misleading information that influence decision making. There is no fact checking mechanism implemented in the models, which worries the participants. Worries about data privacy and protection as well a possible misuse or biased outputs occurred in 14\% of the concern responses. Some participants are worried about the impact on their jobs, especially the possibility of losing their jobs due to AI replacement. Some fear the rapid development of the technology, as both users and the government may not be able to respond adequately.

\section{Discussion and Future Directions}\label{futur}
Over the past, research often focused on logical approaches and large knowledge graphs to deal with commonsense reasoning. Given that we are currently in the era of LLMs which have shown substantial performance improvements across various tasks, we hypothesized that LLMs are capable of handling commonsense reasoning in QA tasks with almost human-level performance (Hypothesis~1). As LLMs are trained on a large number of data and produces human-like text, we assume that it can perform commonsense reasoning without explicit semantic knowledge or logical rules. To show that we evaluated three LLMs on eleven different QA benchmark datasets which are difficult to solve without commonsense reasoning. 

Moreover, we evaluated explanations generated with GPT-3.5 by means on an online questionnaire to investigate how sufficient explanations are to users. Our Hypothesis~2 is stating that an LLM like GPT-3.5 is able to provide good explanations to users without the need of explicit formalized knowledge representation. Most participants are content with GPT-3.5's explanations. Thereby apparently the problem of explainability of AI decisions can be overcome easily.

\subsection{Main Findings}
This study shows that different LLMs reached an overall accuracy of 70.91\% - 89.70\% on eleven challenging QA datasets that are difficult to handle without commonsense reasoning. The new Llama 3 model performed a lot better than Gemma und GPT-3.5 on all eleven commonsense reasoning datasets. 
While there are still challenges such as too little context information, semantic relations or the sciences and medical domain (cf. \cref{anal}), GPT-3.5 still outperforms our questionnaire participants in six out of ten datasets (not considering the medical dataset MedMCQA).  
The results of our questionnaire show that participants answered 69\% of the 20 QA tasks correctly and are outperformed by Llama 3 on every single dataset. We found that Llama 3 is outperforming humans impressively by an average 21\% higher accuracy over ten datasets. Although we only compared the performance of humans versus LLMs on a small amount of examples, we believe that the outcome indicates that our Hypothesis~1 is true and LLMs can handle commonsense reasoning in QA tasks with near-human-level performance, even outperforming human accuracy.

This research focused as well on assessing the explainability of LLMs, recognizing the significant importance of addressing the black-box problem. This is particularly relevant as users need to understand AI decisions. By means of a web-based questionnaire, we evaluated GPT-3.5's explanations for 20 QA tasks. We found a mean linear positive correlation of 0.64 between the comprehensibility of the QA tasks and that of GPT-3.5's explanations. This observation is relevant for the way GPT-3.5's users describe their tasks as it has an impact on the quality of the explanation they receive. 
In our questionnaire, GPT-3.5's explanations were mostly rated \enquote{good} or \enquote{excellent} with 66\% and fair or better with 86\%. Our Hypothesis~2 that LLMs can generate good explanations could be confirmed. However, to improve explanations, it is recommended to not only focus on explaining why one option is correct but also why other answering options are false or less likely.

\subsection{Impact on the Field} 
The development of XAI is facing both scientific and social demands \cite{history-XAI}, and scientists aim to achieve this without sacrificing performance. So far, this grand challenge is mainly dealt with by explicit knowledge, such as knowledge graphs. However, we found that implicit knowledge in the form of probabilistic models can generate good explanations. LLMs made significant advancements in NLP tasks in recent years. Due to the chat function of GPT-3.5, users can easily ask for explanations to understand the response of the AI system. This can tackle the lack of explainability and is a quite simple and yet effective way. Using a questionnaire, we could measure the quality of GPT-3.5's explanations.

Moreover, commonsense reasoning is very important for various NLP tasks. It assesses the relative plausibility of different scenarios and recognizes causality. Until now, research has focused on mathematical logic and the formalization of commonsense reasoning knowledge. However, some philosophers, e.g., Wittgenstein, already claimed that commonsense reasoning knowledge is unformalizable or mathematical logic is inappropriate \cite{McC89}. As seen in our evaluation, LLMs have a good zero-shot performance on different QA tasks that require commonsense reasoning. Nevertheless, we detected seven problems (cf. \cref{anal}) where some LLMs still have difficulties and further research is necessary, e.g., little context information, comparative reasoning, knowledge in the medical and science domain or semantic relations.

Evaluation of LLMs brings AI closer to making a practical impact in the area of XAI and commonsense reasoning. There are still rich opportunities for novel AI research to further measure the quality of explanations as well as opportunities in tackling difficult commonsense reasoning tasks like CommonsenseQA. In future research, one can also investigate certain domains in more detail e.g. focus on the medical domain to foster the way for medical LLM based applications. We believe LLMs have great potential for various applications in combination with robotics, customer service and even legal. However, we have to make sure to keep the concerns like incorrect answers and possible negative impacts on human skills in mind. Furthermore, while the zero-shot performance of the LLMs was already quite good, the accuracy and explanation quality might improve using few-shot learning and chain-of-thought prompting \cite{wei2022chain, gramopadhye2024few}. 

\subsection{Limitations}
Our study has limitations that need to be acknowledged. The number of survey participants we included was rather small, which limits generalization of our results. 
The average age was 26 years and primarily the participants were university students. In general, more participants with diverse ages and nationalities would help to strengthen the results.
Furthermore the key challenge for explainability is to determine what constitutes a "good" explanation since this is subjective and depends on context. We evaluated explanations using a five-level Likert scale from "very poor" to "excellent". However, we only analyzed 20 explanations of GPT-3.5 and argue that our Hypothesis~2 (that LLMs can generate good explanations) can be confirmed. Nevertheless, explainability is very important in the medical field, but we did not consider the MedMCQA dataset in our questionnaire due to a supposed lack of participant's knowledge in medicine.

\section{Conclusion} 
The field of AI has made considerable progress towards large-scale models, especially for NLP tasks. Although the field requires more testing, we argue that LLMs can be used for commonsense reasoning tasks and as well generate helpful explanations for users to understand AI decisions. The use of LLMs is a promising area of research that offers many opportunities to enhance explainability. 
However, to unleash their full potential for XAI, it is crucial to approach the use of these models with caution and to critically evaluate users concerns. 
The explainability of LLMs is crucial for many applications and was an important part of this research. 
We have shown important future directions research involving XAI and commonsense reasoning. Several state-of-the-art LLMs have proven capable of outperforming human performance on a large number of different QA datasets which require commonsense reasoning.

Despite the potential of the field of LLMs, important questions remain for a comprehensive evaluation of LLM explanations. As these key issues are systematically addressed, the potential real world applications will grow. In particular, the stochastic aspects of LLMs, where repeated queries may lead to different answers, should be considered in future work. This would also allow for a better assessment of the error in the LLM performance estimates.

\subsection{Acknowledgments}
We would like to thank all questionnaire participants as well as Osama Siddiqui, Muhammad Hassan Salam and Syed Muhammad Aun Raza Zaidi for their help testing the LLMs on different datasets. 

\bibliography{sample}
\bibliographystyle{splncs04}

%
%
%
\end{document}